\documentclass[11pt]{article}

\usepackage[preprint]{acl}

\usepackage{times}
\usepackage{latexsym}

\usepackage[T1]{fontenc}

\usepackage[utf8]{inputenc}

\usepackage{microtype}

\usepackage{inconsolata}

\usepackage{graphicx}
\usepackage{subcaption}
\usepackage{booktabs} 
\usepackage{amsmath}
\usepackage{amssymb}
\usepackage{mathtools}
\usepackage{amsthm}
\usepackage{multirow}
\usepackage{afterpage}
\usepackage{stfloats}
\usepackage{balance}

\usepackage{xcolor, soul}
\sethlcolor{pink}

\title{Entropy Alone is Insufficient for Safe Selective Prediction in LLMs} 

\author{
 \textbf{Edward Phillips\textsuperscript{1}},
 \textbf{Fredrik~K.~Gustafsson\textsuperscript{1}},
 \textbf{Sean Wu\textsuperscript{1}},
 \textbf{Anshul Thakur\textsuperscript{1}},
 \textbf{David A. Clifton\textsuperscript{1, 2}}
\\
\\
 \textsuperscript{1}Department of Engineering Science, University of Oxford\\
 \textsuperscript{2}Oxford Suzhou Centre for Advanced Research, University of Oxford, Suzhou\\
 \small{
   \textbf{Correspondence:} \href{mailto:edward.phillips@eng.ox.ac.uk}{edward.phillips@eng.ox.ac.uk}
 }
}

\begin{document}
\maketitle

\begin{abstract}
Selective prediction systems can mitigate harms resulting from language model hallucinations by abstaining from answering in high-risk cases. Uncertainty quantification techniques are often employed to identify such cases, but are rarely evaluated in the context of the wider selective prediction policy and its ability to operate at low target error rates. We identify a model-dependent failure mode of entropy-based uncertainty methods that leads to unreliable abstention behaviour, and address it by combining entropy scores with a correctness probe signal. We find that across three QA benchmarks (TriviaQA, BioASQ, MedicalQA) and four model families, the combined score generally improves both the risk--coverage trade-off and calibration performance relative to entropy-only baselines. Our results highlight the importance of deployment-facing evaluation of uncertainty methods, using metrics that directly reflect whether a system can be trusted to operate at a stated risk level.

\end{abstract}

\section{Introduction}

Large language models (LLMs) continue to progress rapidly, having already achieved gold-medal status in maths and coding Olympiads \citep{castelvecchi_deepmind_2025} and written novel peer-reviewed research \citep{yamada2025aiscientistv2workshoplevelautomated}. Despite these advances, even the most capable models produce hallucinations, fluent but factually incorrect answers that are often expressed with a high degree of confidence \citep{huangsurvey2025}. \textit{Selective prediction} systems, which abstain from answering in cases of high uncertainty, can mitigate the harms of propagating such errors in high-stakes domains such as healthcare~\citep{zhou2025hademif}.

Uncertainty quantification (UQ) techniques have been proposed to detect LLM hallucinations, which we categorize broadly into two types: \textit{entropy-based} methods, which treat distributional sharpness over model outputs or activations as a proxy for confidence~\citep{malinin2021uncertainty, chen2024inside, farquhar_detecting_2024}, and \textit{correctness-based} methods, which learn a supervised factuality signal from model activations~\citep{kadavath2022languagemodelsmostlyknow,obeso2025realtimedetectionhallucinatedentities, su2024unsupervisedrealtimehallucinationdetection, han-etal-2025-simple}. These two method types have been found to be only moderately correlated, such that simple combinations can outperform individual methods as hallucination detectors~\citep{xiong2024efficient}. 

Prior work however evaluated such combinations 
primarily via global detection metrics such as AUROC, which can obscure failures that matter most in deployment. We further assess entropy-based UQ methods and their combination with correctness probes through the lens of selective prediction, using deployment-facing metrics: Area Under the Risk--Coverage Curve (AURC)~\citep{zhou2025a} and Target Calibration Error (TCE), defined in Section~\ref{sec:methods}. We include four model families in our experiments, and assess their performance on both general knowledge and medical datasets, where trustworthy systems are of particular importance. We summarize our contributions as follows:

\begin{itemize}
    \item We show that the effectiveness of entropy-based UQ methods is highly model-dependent and identify a characteristic failure mode, the \textit{confidently wrong} regime, in which models produce low-entropy hallucinations.
    \item We show this failure mode is consequential for safe selective prediction, and that augmenting entropy signals with a correctness probe generally improves both hallucination detection and selective prediction performance.
    \item We argue for deployment-facing evaluation of UQ methods: strong AUROC does not imply reliable selective prediction at strict safety thresholds, and metrics such as AURC and TCE better reflect whether a system can be trusted to operate at a stated risk level.
\end{itemize}

\section{Methods}
\label{sec:methods}

\subsection{Selective Prediction Setup}
Given an input question $x$ and model answer $\hat{y}$, let $\mathcal{L}(x)\in\{0,1\}$ denote a binary hallucination indicator for $\hat{y}$ (1 = incorrect). A selective prediction system returns the answer only when a scalar \emph{risk score} $r(x)$ is below a threshold:
\begin{equation}
A_\tau(x)=\mathbb{I}\{r(x)\le \tau\},
\end{equation}
where $A_\tau(x)=1$ means \textsc{Answer} and $A_\tau(x)=0$ means \textsc{Abstain}. Given a target risk $\alpha$, we tune the acceptance threshold $\tau$ on a held-out calibration set. Specifically, we pick the largest $\tau^\star$ for which the observed hallucination rate among answered examples is at most $\alpha$.

\subsection{Uncertainty Signals as Risk Scores}

\paragraph{Entropy-based methods.} We evaluate three entropy-based risk scores. \textbf{Sequence NLL} uses the mean negative log-likelihood of the generated answer, requiring no additional inference beyond the original forward pass. \textbf{Semantic Entropy (SE)} 
measures semantic dispersion by clustering $K$ sampled 
completions through mutual entailment assessment~\citep{farquhar_detecting_2024}. \textbf{Semantic Entropy Probe (SEP)} is a lightweight linear probe trained on prompt representations to predict a binarized SE target, amortizing SE's sampling and entailment cost at test time~\citep{kossen2024semanticentropyprobesrobust}. We convert each entropy-based signal to a risk proxy $u_{\mathrm{ent}}(x)$ such that higher values indicate greater risk.

\paragraph{Probability of Correctness (PC) probe.} Following 
\citet{kadavath2022languagemodelsmostlyknow}, we train a logistic regression classifier on prompt representations to predict whether the model's answer is correct, using externally provided correctness labels. We denote the predicted probability of correctness $p_{\mathrm{pc}}(x) \in [0,1]$ and convert to a risk proxy $u_{\mathrm{pc}}(x) = 1 - p_{\mathrm{pc}}(x)$.

\paragraph{Combination.} Our combined risk score is a two-feature logistic regression over $u_{\mathrm{ent}}(x)$ and $u_{\mathrm{pc}}(x)$, trained on a held-out training split, yielding a single scalar that can be thresholded for 
abstention. We consider an alternative combination in Appendix~\ref{sec:app-combiner-ablation}.

\subsection{Evaluation Metrics}
We report: \textbf{AUROC} and \textbf{AUPRC} for hallucination detection, 
and \textbf{Excess Area Under the Risk--Coverage Curve (E-AURC)} and \textbf{Target Calibration Error (TCE)} to evaluate selective prediction.

\paragraph{E-AURC.}
We form a risk--coverage (RC) curve by sorting examples by the risk score and, for each acceptance threshold, measuring (a) \emph{coverage}, the fraction of examples answered, and (b) \emph{selective risk}, the hallucination rate among answered examples. AURC is the area under this RC curve. E-AURC is the excess area relative to an oracle ranking that abstains on all hallucinations before abstaining on any correct examples, providing a unitless measure in $[0, 1]$~\citep{geifman2018biasreduced}.

\paragraph{TCE.}
For calibration, we sweep target risks $\alpha$ over an operational range $\mathcal{A}$ and measure how closely the calibrated policy achieves each target on the test set:
\[
\mathrm{TCE}=\mathbb{E}_{\alpha\sim\mathcal{A}}\big[|R_{\text{test}}(\alpha)-\alpha|\big].
\]
where $R_{\text{test}}(\alpha)$ is the test set selective risk when the acceptance threshold is tuned to target $\alpha$.

We use TCE rather than standard calibration metrics such as ECE because our goal is accurate risk control at \emph{specific operating points}, rather than global probability calibration over $[0,1]$.

\begin{figure*}[h]
    \centering
    \includegraphics[width=\linewidth]{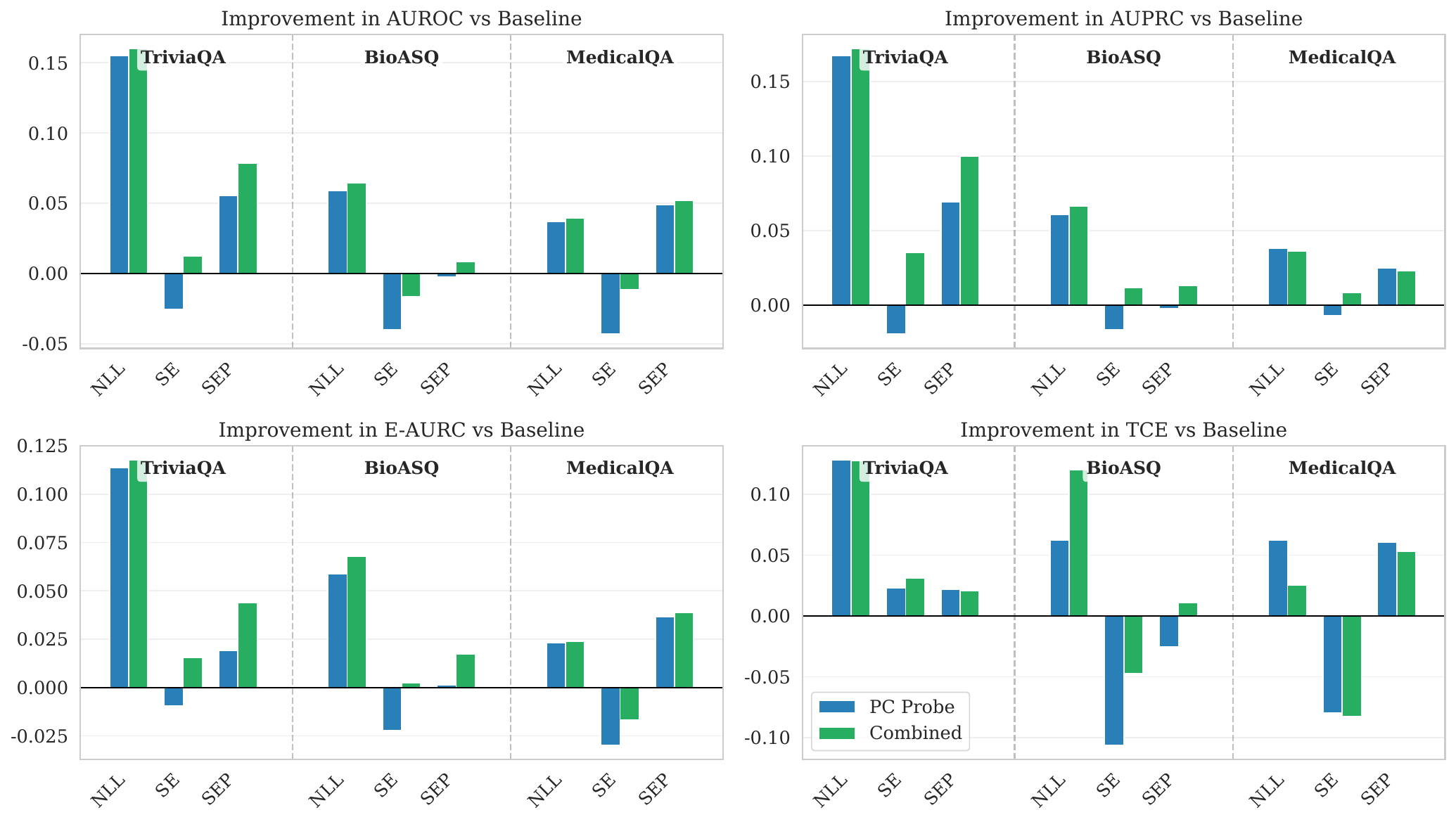}
    \caption{Performance of the PC probe and combined method relative to entropy-only baselines, averaged across all four evaluated models. Positive values denote an improvement in the target metric (i.e., higher AUROC/AUPRC or lower E-AURC/TCE). Abbreviations: NLL (Sequence NLL), SE (Semantic Entropy), SEP (Semantic Entropy Probe). The combined method consistently outperforms both the entropy baselines and the standalone PC probe across most configurations, with the notable exception of SE on the MedicalQA dataset.}
    \label{fig:solution_bars}
\end{figure*}

\section{Experiments}

\subsection{Models and Datasets}
\label{subsec:models-datasets}

We consider three QA benchmarks:
\textbf{TriviaQA}, consisting of open-domain factoid questions with short reference answers~\citep{joshi-etal-2017-triviaqa}; \textbf{BioASQ}, comprising biomedical questions grounded in PubMed abstracts~\citep{tsatsaronis2015overview}; and \textbf{MedicalQA}, an open-ended medical QA benchmark converted from multiple-choice questions in MedQA and MedMCQA~\citep{chen-etal-2025-towards-medical}. 

We evaluate across four model families: \textbf{Mistral} (Ministral-8B-Instruct-2410), \textbf{Llama} (Llama-3.2-3B-Instruct), \textbf{Qwen} (Qwen3-4B-Instruct-2507), and \textbf{Gemma} (Gemma-3-4B)~\citep{grattafiori2024llama, mistralai_ministraux_2024, yang2025qwen3technicalreport, gemmateam2025gemma3technicalreport}. We provide further details of dataset generation and probe supervision in Appendix~\ref{sec:app-method-details}. 

For TCE, we set $\mathcal{A} = [0.05, 0.30]$: the lower bound reflects the difficulty of these benchmarks at the evaluated model scales, and the upper bound approximates the base hallucination rate, above which abstention yields no benefit.

\subsection{Model-Specific Failure Modes}

\begin{figure}[h]
    \centering
    \includegraphics[width=\linewidth]{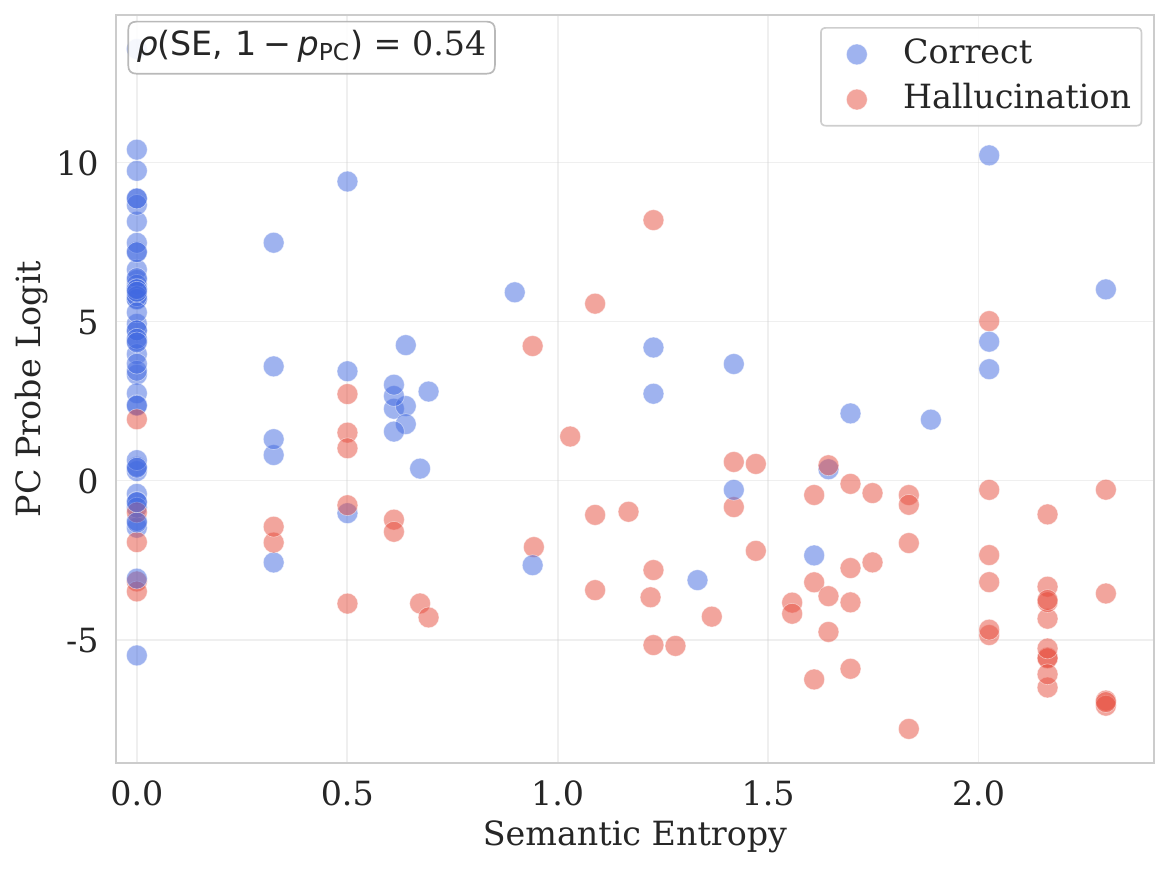}
    \caption{SE score vs.\ PC probe logit for Qwen on TriviaQA. Many answers, both correct and hallucinated, collapse to $\mathrm{SE}=0$, yet receive distinct PC probe scores.}
    \label{fig:qwen-scatter}
\end{figure}

\begin{table}
\centering
\small
\caption{Hallucination detection AUROC for SE and PC probe on TriviaQA, with Spearman correlation $\rho_s$ between their risk scores. The best method varies across models, with only moderate correlation between scores.}\vspace{-2.0mm}
\label{tab:corr}
\begin{tabular}{lccc}
\toprule
\textbf{Model} & \textbf{SE AUROC} & \textbf{PC AUROC} & $\rho_s$ \\
\midrule
Llama 3.2 3B & $\mathbf{0.884}$ & $0.813$ & 0.51 \\
Qwen3 4B & $0.844$ & $\mathbf{0.888}$ & 0.54 \\
Gemma 3 4B & $\mathbf{0.752}$ & $\mathbf{0.752}$ & 0.34 \\
Ministral 8B & $\mathbf{0.823}$ & $0.750$ & 0.50 \\
\bottomrule
\end{tabular}
\end{table}

Previous work found entropy-based methods to outperform correctness probe-based methods on knowledge-seeking tasks, such as TriviaQA~\citep{xiong2024efficient}; we find that UQ method efficacy is also highly model-dependent. Table \ref{tab:corr} shows that SE is the stronger detector for Llama and Ministral on TriviaQA, while the PC probe 
dominates for Qwen. Across all models, the two signals are only moderately correlated, suggesting they capture 
complementary aspects of model uncertainty.

Figure~\ref{fig:qwen-scatter} illustrates the primary failure mode of SE for the Qwen model. A substantial fraction of questions are assigned near-zero semantic entropy, collapsing to a vertical line. This cluster includes both correct answers and hallucinations, the latter being \textit{confidently wrong} instances in which the output distribution lacks diversity despite being incorrect~\citep{phillips2025geometricuncertaintydetectingcorrecting}. The PC probe does not exhibit this degeneracy: questions assigned identical zero SE scores are spread across a range of PC probe logits, with hallucinations often receiving elevated risk scores. This observation, together with the moderate Spearman correlation, motivates combining the two signals for selective prediction.

\subsection{Combining Entropy and Correctness}

\begin{figure*}[h]
    \centering
    \includegraphics[width=\linewidth]{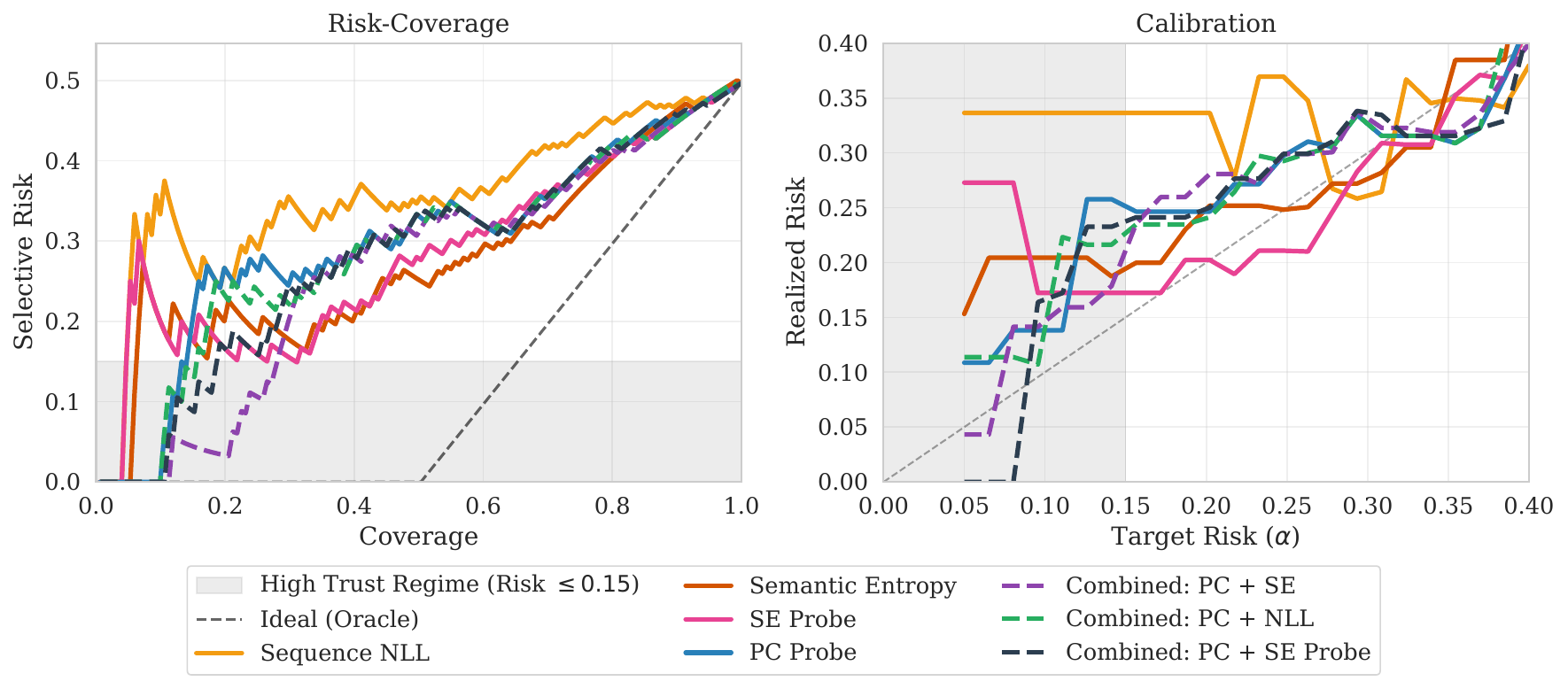}
    \caption{Ministral 8B selective prediction on BioASQ. \textit{Left:} risk--coverage curves; the shaded region marks a high-trust regime ($\alpha \leq 0.15$). Entropy based methods fail to enter this regime at non-trivial coverage. \textit{Right:} realized hallucination rate vs.\ target $\alpha$; a perfect system lies on the diagonal. Entropy-based methods diverge sharply at strict targets; combined methods do not.}
    \label{fig:ministral-bioasq}
\end{figure*}

Figure~\ref{fig:solution_bars} shows the effect of adding correctness signals to each entropy-based baseline, averaged across all four models. Across all three datasets, the combined method (entropy + PC probe) produces consistent improvements in AUROC and AUPRC over the entropy-only baselines. Improvements in the deployment-facing metrics are similarly broad: E-AURC and TCE both decrease for the large majority of (method, dataset) combinations, indicating a better risk--coverage trade-off and more reliable calibration to specified targets.

Using the PC probe alone as a replacement for entropy also improves on the baseline in many cases, but the combined score generally yields the largest and most consistent gains, suggesting the two signals carry complementary information.

The clearest exception is SE on MedicalQA, where neither the PC probe alone nor the combination reliably improves TCE. This is consistent with MedicalQA being the hardest benchmark in our evaluation: when base model accuracy is low, the correctness probe itself becomes harder to calibrate, limiting the benefit of combination.

\subsection{Correctness Signals Can Be Effective in Strict Safety Regimes}

Figure~\ref{fig:ministral-bioasq} illustrates a practical consequence of poor discrimination among confident samples for a deployed selective prediction system, shown for the Ministral 8B model on the BioASQ dataset. Entropy-based methods exhibit a \emph{risk floor} where their RC curves fail to enter a high-trust regime at non-trivial coverage. This directly causes the calibration failure shown in the right panel: when a threshold is tuned to a strict target $\alpha$, the realized hallucination rate remains well above it. Methods incorporating the PC probe lower the risk floor, and consequently track the calibration diagonal more closely throughout the high-trust regime.

\section{Conclusion}

Selective prediction offers a principled way to mitigate hallucination harms in LLMs, but our results show that entropy-based uncertainty signals alone are often insufficient for reliable operation at strict safety targets. We identified a characteristic failure mode, the \textit{confidently wrong} regime, in which models produce low-entropy hallucinations. This failure mode is model-dependent and has consequences for safe selective prediction.

We showed that a lightweight combination of entropy-based scores with a correctness probe addresses this failure mode, consistently improving both the risk–coverage trade-off (E-AURC) and calibration to specified targets (TCE) across three QA benchmarks and four model families. These gains are achieved without additional inference overhead beyond a single forward pass of the LLM.

Our findings argue for a shift in how UQ methods are evaluated: global detection metrics should be complemented by deployment-facing selective prediction criteria, particularly TCE, which directly reflects whether a system can be trusted to operate at a stated risk level.

\newpage

\section*{Limitations}

Our experiments are limited to short-form QA tasks, where answers can be judged as hallucination relative to reference strings. It is unclear whether the confidently wrong failure mode, or the complementarity between entropy and correctness probe signals, generalizes to long-form generation, multi-step reasoning, or open-ended tasks where hallucinations can be harder to define.

We evaluate a relatively small set of model families in the 3B–8B parameter range. The degree to which entropy-based signals degrade (particularly the pattern observed with Qwen) may vary substantially across architectures, scales, and training recipes, and larger frontier models may exhibit different failure profiles.

Our correctness probe requires supervised training on labelled examples, introducing an annotation dependency that may limit applicability in low-resource settings. Furthermore, the combiner is trained on the same split used to fit the PC probe, meaning it sees probe predictions that were not held out; this introduces a mild optimistic bias, mitigated but not eliminated by strong regularization. Future work should explore whether similar complementarity can be achieved with unsupervised or self-supervised correctness proxies.

Finally, our TCE estimates fall back to base hallucination rate when coverage is zero. This penalizes complete abstention less harshly than some alternative designs, and results at very strict targets should be interpreted with this in mind.

\section*{Acknowledgments}

DAC was funded by an NIHR Research Professorship; a Royal Academy of Engineering Research Chair; and the InnoHK Hong Kong Centre for Cerebro-cardiovascular Engineering (COCHE); and was supported by the National Institute for Health Research (NIHR) Oxford Biomedical Research Centre (BRC) and the Pandemic Sciences Institute at the University of Oxford. EP was funded by an NIHR Research Studentship. SW was supported by the Rhodes Scholarship.

\bibliography{custom}

\renewcommand{\thefigure}{A\arabic{figure}}
\setcounter{figure}{0}
\renewcommand{\thetable}{A\arabic{table}}
\setcounter{table}{0}
\renewcommand{\theequation}{A\arabic{equation}}
\setcounter{equation}{0}

\newpage


\appendix

\twocolumn[{%
  \begin{center}
    \LARGE \textbf{Appendices}
    \vspace{2em} 
  \end{center}
}]

\section{Method Details}
\label{sec:app-method-details}

\paragraph{Dataset generation.}
For each dataset, we compile a question set of 1000 examples using a fixed random seed, shared across all models. For each question and model, we generate one low-temperature answer $\hat{y}$ at $T=0.1$, used for correctness judging, and $K=10$ higher-temperature samples at $T=1$, used to compute semantic entropy. Correctness labels are assigned via an LLM-as-judge assessment following the pipeline of \citet{kossen2024semanticentropyprobesrobust}. For each $(\text{model},\text{dataset})$ pair, examples are split deterministically into train, calibration, and test sets in a 70:15:15 ratio, stratified on that model's correctness label. 

\paragraph{Semantic entropy computation.}
Semantic entropy is computed from the $K$ high-temperature samples using a DeBERTa-based model to assign samples to 
semantic clusters via mutual entailment, following 
\citet{farquhar_detecting_2024}. Specifically, we use the 
\texttt{cluster\_assignment\_entropy} variant, which computes 
entropy over the discrete cluster assignment distribution 
rather than over the token-level log-likelihoods.

\paragraph{Probe architecture and training.}
Both the PC probe and the SEP are logistic regression 
classifiers (L2 regularisation, $C=0.1$) with a 
\texttt{StandardScaler} preprocessing step, implemented via 
\texttt{scikit-learn}. The PC probe is trained to predict 
binary correctness ($\hat{y}$ correct or not); the SEP is 
trained to predict a binarized semantic entropy label 
(above/below the training-split median).

\paragraph{Token position.}
Following \citet{kossen2024semanticentropyprobesrobust}, we 
extract representations at the last token before generation 
begins (TBG), enabling pre-generation abstention. We select 
the best-performing layer at this position per probe via 
5-fold stratified cross-validation on the training split 
only, then refit on all training data at the chosen layer.

\paragraph{Combiner training.}
The combined risk score is a two-feature logistic regression 
trained on the training split, taking as input the entropy 
risk proxy $u_{\mathrm{ent}}(x)$ and the PC probe risk proxy 
$u_{\mathrm{pc}}(x) = 1 - p_{\mathrm{pc}}(x)$. Both the PC 
probe and the combiner are trained on the same split; the 
combiner therefore sees in-sample PC probe predictions during 
training, which is a mild source of optimism mitigated by the 
strong regularization ($C=0.1$) applied to both models.

\section{Further Related Work}

\subsection{Selective Prediction}
In deployment, uncertainty estimates are often used for selective prediction, where the system answers only when a confidence score exceeds a threshold.  \citet{cole-etal-2023-selectively} compared selective prediction methods for QA, finding that sampling-based consistency measures generally outperform likelihood methods, particularly for ambiguous questions. We build on this work by studying a much broader spectrum of models, datasets, and UQ methods, and by investigating specific failure modes and behaviour in strict safety regimes.

Recent work also studies training models to abstain, rather than post-hoc thresholding. \citet{tjandra2024finetuninglargelanguagemodels} fine-tune LLMs using Semantic Entropy signals and propose an \emph{Accuracy-Engagement Distance} metric that jointly captures correctness and willingness to answer. We do not report this metric since it investigates a single operating point in risk--coverage space, whereas we evaluate models over a range of operating points.

\subsection{Risk Control and Conformal Prediction}

Beyond ranking and risk--coverage curves, deploying abstention policies safely requires controlling the hallucination rate among accepted answers. Conformal prediction (CP) \citep{vovk2005algorithmic} provides finite-sample validity guarantees for prediction sets; split conformal methods are particularly practical \citep{scp}. Although CP was originally developed for classification and regression, recent work has adapted conformal ideas to language generation and factuality control.

\citet{mohri2024language} proposed \emph{Conformal Factuality}, selecting the most specific claim in a back-off chain that satisfies a correctness guarantee. Other work addresses the non-exchangeable nature of text streams: \citet{ulmer2024nonexchangeableconformallanguagegeneration} constructed calibrated next-token sets under distribution shift using nearest-neighbor retrieval. In a related direction, \citet{xu2025tecptokenentropyconformalprediction} proposed using token-level entropy as a conformal nonconformity score to calibrate generation-time uncertainty. 

\begin{table*}[!b]
    \centering
    \caption{Performance of base hallucination detection methods across datasets, averaged over all models.}
    \label{tab:app_methods_ind}
\begin{tabular}{ll|cccc}
\toprule
\textbf{Dataset} & \textbf{Method} & \textbf{AUROC} $\uparrow$ & \textbf{AUPRC} $\uparrow$ & \textbf{E-AURC} $\downarrow$ & \textbf{TCE} $\downarrow$ \\
\midrule
\multirow{4}{*}{\textbf{TriviaQA}} & Sequence NLL & $0.646_{11.5}$ & $0.544_{3.4}$ & $0.217_{9.9}$ & $0.189_{11.5}$ \\
 & Semantic Entropy & $\mathbf{0.826_{5.5}}$ & $\mathbf{0.732_{6.7}}$ & $\mathbf{0.094_{5.5}}$ & $0.084_{8.7}$ \\
 & SE Probe & $0.745_{2.5}$ & $0.643_{9.9}$ & $0.122_{1.4}$ & $0.083_{4.3}$ \\
 & PC Probe & $0.801_{6.5}$ & $0.712_{12.9}$ & $0.103_{4.1}$ & $\mathbf{0.061_{1.8}}$ \\
\midrule
\multirow{4}{*}{\textbf{BioASQ}} & Sequence NLL & $0.662_{1.4}$ & $0.694_{4.8}$ & $0.245_{2.6}$ & $0.278_{12.9}$ \\
 & Semantic Entropy & $\mathbf{0.760_{4.2}}$ & $\mathbf{0.771_{5.4}}$ & $\mathbf{0.164_{3.3}}$ & $\mathbf{0.109_{6.0}}$ \\
 & SE Probe & $0.723_{8.0}$ & $0.758_{6.8}$ & $0.187_{6.1}$ & $0.189_{9.5}$ \\
 & PC Probe & $0.721_{4.1}$ & $0.755_{4.9}$ & $0.186_{4.1}$ & $0.215_{14.0}$ \\
\midrule
\multirow{4}{*}{\textbf{MedicalQA}} & Sequence NLL & $0.622_{5.7}$ & $0.835_{3.9}$ & $0.260_{3.0}$ & $0.553_{9.2}$ \\
 & Semantic Entropy & $\mathbf{0.702_{5.2}}$ & $\mathbf{0.880_{2.7}}$ & $\mathbf{0.206_{3.6}}$ & $\mathbf{0.411_{18.2}}$ \\
 & SE Probe & $0.610_{4.8}$ & $0.848_{3.1}$ & $0.273_{4.7}$ & $0.551_{13.3}$ \\
 & PC Probe & $0.659_{7.7}$ & $0.873_{4.1}$ & $0.236_{6.2}$ & $0.491_{11.3}$ \\
\bottomrule
\end{tabular}
\end{table*}

Our setting differs in two ways: (i) we target \emph{binary selective answering} of atomic responses (answer vs.\ abstain) rather than producing calibrated token-level sets or modifying output specificity; and (ii) we draw inspiration from conformal calibration in using a held-out set to tune acceptance thresholds, but we do not claim conformal validity for the resulting selective-risk guarantee (e.g., under distribution shift or non-exchangeability).

\subsection{Alternative Uncertainty Methods}

A common approach to detecting hallucinations is to extract uncertainty estimates directly from the model's output distribution. Variant methods include those using token-level statistics such as maximum softmax probabilities \citep{plaut2024softmax}, or attention-based metrics \citep{fomicheva2020unsupervised}. While computationally cheap, these measures are sensitive to lexical redundancy; a model may split probability mass across semantically identical but lexically distinct phrasings.

To address this, many methods estimate uncertainty at the semantic level. Follow-up work to semantic entropy explored alternative semantic representations and estimators, including geometric perspectives on the semantic space \citep{phillips2025geometricuncertaintydetectingcorrecting, phillips2026semanticselfdistillationlanguagemodel} and density-based measures \citep{qiu2024semantic}. \citet{chen2024inside} proposed EigenScore, which analyzes the covariance eigenvalues of hidden states across multiple samples to measure semantic divergence. 

Rather than exhaustively evaluating all such variants, our main experiments select representative baselines based on the UQ taxonomy outlined by \citet{xiong2024efficient}, ensuring we cover the three operational paradigms:  \textbf{(i) single-sample} (Sequence NLL) for cheap but lexically sensitive metrics; \textbf{(ii) multi-sample} (Semantic Entropy) for robust but expensive dispersion measures; and \textbf{(iii) probe-based} (Semantic Entropy Probe) for supervised, single-pass approximations.

\section{Baseline Method Performance}

Table~\ref{tab:app_methods_ind} reports the performance of each baseline method averaged across models. Semantic Entropy is the strongest individual method on TriviaQA and BioASQ across all four metrics, consistent with prior findings on knowledge-seeking tasks~\citep{farquhar_detecting_2024}. However, SE requires $K$ additional forward passes and an entailment step at test time, making it considerably more expensive than the other baseline methods, which are all single-pass. Among these cheaper methods, the PC probe is the strongest performer on TriviaQA (TCE $0.061$) and competitive on  BioASQ, while SE Probe - which amortizes SE's sampling cost at the expense of supervised training - offers a middle ground. On MedicalQA all methods degrade substantially, reflecting the difficulty of the task.

\clearpage
\onecolumn

\section{Full Hallucination Detection Results}

\begin{table*}[!h]
\centering
\caption{Detailed hallucination detection classification metrics (AUROC and AUPRC) for all evaluated models and datasets. Results are reported as $\text{Mean}_{\text{Std}}$ across $200$ bootstrap iterations. The best performing method per column is highlighted in bold, and \textit{Acc} denotes the base model accuracy at full coverage.}
\label{tab:app_detection}
\begin{tabular}{lcccccc}
\toprule
\textbf{Method} & \multicolumn{2}{c}{\textbf{TriviaQA}} & \multicolumn{2}{c}{\textbf{BioASQ}} & \multicolumn{2}{c}{\textbf{MedicalQA}} \\
 & \scriptsize{AUROC} & \scriptsize{AUPRC} & \scriptsize{AUROC} & \scriptsize{AUPRC} & \scriptsize{AUROC} & \scriptsize{AUPRC} \\
\midrule
\textbf{Llama 3.2 3B} & \multicolumn{2}{c}{\scriptsize{\textit{Acc: 64.2\%}}} & \multicolumn{2}{c}{\scriptsize{\textit{Acc: 48.3\%}}} & \multicolumn{2}{c}{\scriptsize{\textit{Acc: 22.0\%}}} \\
Sequence NLL & $0.664_{4.5}$ & $0.531_{7.2}$ & $0.656_{4.0}$ & $0.633_{5.5}$ & $\mathbf{0.690_{5.5}}$ & $0.853_{4.3}$ \\
Semantic Entropy & $\mathbf{0.884_{2.7}}$ & $0.767_{5.8}$ & $\mathbf{0.737_{4.2}}$ & $0.699_{5.8}$ & $0.688_{5.6}$ & $0.866_{3.4}$ \\
SE Probe & $0.723_{4.4}$ & $0.606_{6.8}$ & $0.697_{4.3}$ & $0.721_{5.4}$ & $0.626_{5.4}$ & $0.864_{3.2}$ \\
PC Probe & $0.813_{3.5}$ & $0.691_{6.6}$ & $0.669_{3.6}$ & $0.683_{5.4}$ & $0.642_{5.3}$ & $0.868_{3.3}$ \\
Combined: PC + NLL & $0.826_{3.4}$ & $0.715_{6.2}$ & $0.674_{3.6}$ & $0.685_{5.2}$ & $0.649_{5.2}$ & $\mathbf{0.877_{3.0}}$ \\
Combined: PC + SE & $0.863_{2.9}$ & $\mathbf{0.776_{5.4}}$ & $0.702_{3.5}$ & $\mathbf{0.725_{5.0}}$ & $0.666_{5.2}$ & $0.875_{3.3}$ \\
Combined: PC + SE Probe & $0.836_{3.2}$ & $0.744_{5.8}$ & $0.693_{3.6}$ & $0.713_{4.9}$ & $0.654_{5.4}$ & $0.870_{3.4}$ \\
\addlinespace
\textbf{Qwen3 4B} & \multicolumn{2}{c}{\scriptsize{\textit{Acc: 52.3\%}}} & \multicolumn{2}{c}{\scriptsize{\textit{Acc: 40.4\%}}} & \multicolumn{2}{c}{\scriptsize{\textit{Acc: 23.3\%}}} \\
Sequence NLL & $0.632_{3.7}$ & $0.580_{5.5}$ & $0.663_{4.1}$ & $0.728_{4.6}$ & $0.613_{4.4}$ & $0.815_{4.1}$ \\
Semantic Entropy & $0.844_{3.2}$ & $0.800_{5.0}$ & $0.717_{3.9}$ & $0.765_{4.6}$ & $0.648_{5.2}$ & $0.850_{3.9}$ \\
SE Probe & $0.779_{3.9}$ & $0.769_{4.5}$ & $0.622_{4.4}$ & $0.681_{5.3}$ & $0.643_{5.3}$ & $0.852_{4.3}$ \\
PC Probe & $0.888_{2.5}$ & $0.883_{3.2}$ & $0.709_{4.3}$ & $0.781_{4.4}$ & $0.671_{4.8}$ & $\mathbf{0.884_{3.2}}$ \\
Combined: PC + NLL & $0.888_{2.6}$ & $0.865_{4.5}$ & $0.711_{4.3}$ & $0.771_{4.9}$ & $0.667_{5.1}$ & $0.861_{4.3}$ \\
Combined: PC + SE & $\mathbf{0.910_{2.1}}$ & $\mathbf{0.910_{2.4}}$ & $\mathbf{0.722_{4.1}}$ & $\mathbf{0.790_{4.4}}$ & $\mathbf{0.681_{4.8}}$ & $0.876_{3.8}$ \\
Combined: PC + SE Probe & $0.894_{2.4}$ & $0.884_{3.4}$ & $0.707_{4.3}$ & $0.773_{4.5}$ & $0.680_{4.8}$ & $0.882_{3.6}$ \\
\addlinespace
\textbf{Gemma 3 4B} & \multicolumn{2}{c}{\scriptsize{\textit{Acc: 55.6\%}}} & \multicolumn{2}{c}{\scriptsize{\textit{Acc: 36.4\%}}} & \multicolumn{2}{c}{\scriptsize{\textit{Acc: 17.3\%}}} \\
Sequence NLL & $0.504_{2.9}$ & $0.503_{4.1}$ & $0.647_{4.1}$ & $0.737_{4.0}$ & $0.635_{3.3}$ & $0.880_{2.5}$ \\
Semantic Entropy & $0.752_{4.1}$ & $0.712_{5.3}$ & $\mathbf{0.777_{3.9}}$ & $0.824_{4.2}$ & $0.701_{5.7}$ & $0.905_{3.0}$ \\
SE Probe & $0.748_{3.9}$ & $0.663_{6.2}$ & $0.776_{4.2}$ & $\mathbf{0.825_{4.1}}$ & $0.634_{6.3}$ & $0.873_{4.3}$ \\
PC Probe & $0.752_{4.5}$ & $0.704_{6.3}$ & $0.746_{4.5}$ & $0.788_{4.7}$ & $0.755_{5.9}$ & $0.921_{3.2}$ \\
Combined: PC + NLL & $0.756_{4.5}$ & $0.714_{6.1}$ & $0.751_{4.4}$ & $0.815_{4.0}$ & $0.770_{5.4}$ & $0.934_{2.6}$ \\
Combined: PC + SE & $\mathbf{0.785_{4.2}}$ & $\mathbf{0.762_{5.1}}$ & $0.761_{4.3}$ & $0.823_{3.9}$ & $\mathbf{0.789_{5.0}}$ & $\mathbf{0.944_{2.2}}$ \\
Combined: PC + SE Probe & $0.776_{4.2}$ & $0.751_{5.3}$ & $0.753_{4.4}$ & $0.817_{3.8}$ & $0.749_{6.1}$ & $0.915_{3.4}$ \\
\addlinespace
\textbf{Ministral 8B} & \multicolumn{2}{c}{\scriptsize{\textit{Acc: 71.5\%}}} & \multicolumn{2}{c}{\scriptsize{\textit{Acc: 50.3\%}}} & \multicolumn{2}{c}{\scriptsize{\textit{Acc: 23.3\%}}} \\
Sequence NLL & $0.783_{4.2}$ & $0.564_{8.2}$ & $0.681_{4.3}$ & $0.678_{5.6}$ & $0.551_{5.9}$ & $0.793_{4.5}$ \\
Semantic Entropy & $\mathbf{0.823_{3.6}}$ & $\mathbf{0.648_{7.1}}$ & $\mathbf{0.811_{3.6}}$ & $0.797_{4.7}$ & $\mathbf{0.772_{4.2}}$ & $\mathbf{0.901_{2.7}}$ \\
SE Probe & $0.729_{4.5}$ & $0.534_{7.5}$ & $0.798_{3.8}$ & $\mathbf{0.803_{4.5}}$ & $0.539_{5.2}$ & $0.804_{4.2}$ \\
PC Probe & $0.750_{4.4}$ & $0.570_{7.1}$ & $0.759_{3.9}$ & $0.767_{4.9}$ & $0.569_{5.1}$ & $0.821_{3.8}$ \\
Combined: PC + NLL & $0.753_{4.3}$ & $0.573_{7.1}$ & $0.767_{3.8}$ & $0.770_{5.1}$ & $0.560_{5.3}$ & $0.814_{3.9}$ \\
Combined: PC + SE & $0.796_{3.8}$ & $0.620_{7.0}$ & $0.791_{3.6}$ & $0.794_{4.6}$ & $0.628_{4.9}$ & $0.859_{3.0}$ \\
Combined: PC + SE Probe & $0.788_{3.7}$ & $0.593_{6.9}$ & $0.774_{3.7}$ & $0.778_{4.6}$ & $0.565_{5.2}$ & $0.818_{3.8}$ \\
\addlinespace
\bottomrule
\end{tabular}
\end{table*}

\clearpage
\section{Full Selective Prediction Results}

\begin{table*}[!h]
    \caption{Detailed selective prediction metrics: Full-coverage Excess Area Under the Risk--Coverage Curve (E-AURC) and Target Calibration Error (TCE) for all evaluated models and datasets. TCE is averaged over operational target risk levels $\alpha \in[0.05, 0.3]$. Lower values indicate better performance, with the best method per column highlighted in bold. In cases where coverage is zero over the whole operational range, we report TCE as the model base hallucination rate.}
    \label{tab:sp_results}
\begin{tabular}{lcccccc}
\toprule
\textbf{Method} & \multicolumn{2}{c}{\textbf{TriviaQA}} & \multicolumn{2}{c}{\textbf{BioASQ}} & \multicolumn{2}{c}{\textbf{MedicalQA}} \\
 & \scriptsize{E-AURC} $\downarrow$ & \scriptsize{TCE} $\downarrow$ & \scriptsize{E-AURC} $\downarrow$ & \scriptsize{TCE} $\downarrow$ & \scriptsize{E-AURC} $\downarrow$ & \scriptsize{TCE} $\downarrow$ \\
\midrule
\multicolumn{7}{l}{\textbf{Llama 3.2 3B}} \\
Sequence NLL & $0.194_{4.2}$ & $0.181_{8.1}$ & $0.244_{3.9}$ & $0.198_{7.6}$ & $0.225_{3.4}$ & $0.437_{7.7}$ \\
Semantic Entropy & $\mathbf{0.045_{1.3}}$ & $\mathbf{0.041_{1.2}}$ & $\mathbf{0.163_{3.2}}$ & $\mathbf{0.073_{1.2}}$ & $\mathbf{0.210_{3.7}}$ & $\mathbf{0.332_{9.7}}$ \\
SE Probe & $0.119_{2.5}$ & $0.073_{4.8}$ & $0.214_{3.9}$ & $0.294_{15.1}$ & $0.266_{3.6}$ & $0.357_{1.3}$ \\
PC Probe & $0.087_{2.2}$ & $0.073_{4.3}$ & $0.237_{3.3}$ & $0.263_{16.8}$ & $0.242_{3.6}$ & $0.403_{4.1}$ \\
Combined: PC + NLL & $0.077_{2.0}$ & $0.080_{4.8}$ & $0.217_{3.1}$ & $0.166_{6.2}$ & $0.240_{3.5}$ & $0.405_{3.8}$ \\
Combined: PC + SE & $0.055_{1.5}$ & $0.049_{1.2}$ & $0.184_{2.8}$ & $0.095_{2.1}$ & $0.236_{3.5}$ & $0.414_{5.0}$ \\
Combined: PC + SE Probe & $0.070_{1.7}$ & $0.103_{10.4}$ & $0.194_{2.9}$ & $0.151_{14.5}$ & $0.237_{3.6}$ & $0.438_{4.2}$ \\
\addlinespace
\multicolumn{7}{l}{\textbf{Qwen3 4B}} \\
Sequence NLL & $0.259_{4.3}$ & $0.267_{12.1}$ & $0.251_{3.7}$ & $0.260_{7.7}$ & $0.280_{3.6}$ & $0.536_{7.7}$ \\
Semantic Entropy & $0.094_{2.7}$ & $\mathbf{0.045_{0.9}}$ & $0.210_{3.5}$ & $\mathbf{0.197_{7.0}}$ & $0.248_{3.4}$ & $0.598_{7.7}$ \\
SE Probe & $0.141_{3.4}$ & $0.082_{4.6}$ & $0.260_{3.6}$ & $0.245_{8.5}$ & $0.265_{3.6}$ & $0.598_{7.7}$ \\
PC Probe & $0.066_{1.8}$ & $0.063_{2.0}$ & $0.201_{3.6}$ & $0.389_{13.8}$ & $0.249_{3.5}$ & $0.459_{15.9}$ \\
Combined: PC + NLL & $0.065_{1.9}$ & $0.062_{2.7}$ & $\mathbf{0.192_{3.5}}$ & $0.300_{18.7}$ & $0.249_{3.5}$ & $0.458_{16.0}$ \\
Combined: PC + SE & $\mathbf{0.046_{1.2}}$ & $0.077_{3.2}$ & $0.202_{3.7}$ & $0.389_{13.7}$ & $\mathbf{0.243_{3.4}}$ & $\mathbf{0.374_{6.2}}$ \\
Combined: PC + SE Probe & $0.057_{1.5}$ & $0.064_{2.5}$ & $0.199_{3.5}$ & $0.387_{14.1}$ & $0.246_{3.4}$ & $0.452_{16.3}$ \\
\addlinespace
\multicolumn{7}{l}{\textbf{Gemma 3 4B}} \\
Sequence NLL & $0.323_{4.2}$ & $0.280_{7.7}$ & $0.274_{4.1}$ & $0.465_{7.7}$ & $0.245_{3.3}$ & $0.657_{7.7}$ \\
Semantic Entropy & $0.170_{3.6}$ & $0.215_{14.9}$ & $0.152_{3.2}$ & $0.094_{3.3}$ & $0.208_{3.5}$ & $\mathbf{0.519_{7.7}}$ \\
SE Probe & $0.123_{2.3}$ & $0.141_{3.9}$ & $\mathbf{0.142_{3.1}}$ & $0.120_{5.3}$ & $0.224_{3.6}$ & $0.657_{7.7}$ \\
PC Probe & $0.161_{3.6}$ & $0.074_{3.8}$ & $0.162_{3.1}$ & $0.136_{8.8}$ & $0.152_{3.3}$ & $0.657_{7.7}$ \\
Combined: PC + NLL & $0.158_{3.6}$ & $0.072_{4.3}$ & $0.162_{3.1}$ & $0.095_{3.8}$ & $0.149_{3.2}$ & $0.657_{7.7}$ \\
Combined: PC + SE & $0.138_{3.5}$ & $0.057_{1.5}$ & $0.148_{2.9}$ & $\mathbf{0.079_{1.0}}$ & $\mathbf{0.144_{3.1}}$ & $0.657_{7.7}$ \\
Combined: PC + SE Probe & $\mathbf{0.119_{2.5}}$ & $\mathbf{0.052_{1.2}}$ & $0.158_{3.1}$ & $0.108_{4.7}$ & $0.153_{3.3}$ & $0.657_{7.7}$ \\
\addlinespace
\multicolumn{7}{l}{\textbf{Ministral 8B}} \\
Sequence NLL & $0.091_{2.5}$ & $0.031_{0.4}$ & $0.212_{3.5}$ & $0.187_{6.8}$ & $0.289_{3.8}$ & $0.583_{10.1}$ \\
Semantic Entropy & $\mathbf{0.066_{1.9}}$ & $0.036_{0.8}$ & $0.132_{3.3}$ & $0.071_{3.4}$ & $\mathbf{0.161_{3.1}}$ & $\mathbf{0.196_{2.0}}$ \\
SE Probe & $0.107_{2.6}$ & $0.038_{0.9}$ & $0.133_{3.4}$ & $0.100_{6.1}$ & $0.337_{3.1}$ & $0.593_{7.7}$ \\
PC Probe & $0.099_{2.4}$ & $0.035_{0.8}$ & $0.146_{2.8}$ & $0.073_{2.2}$ & $0.303_{3.6}$ & $0.445_{4.3}$ \\
Combined: PC + NLL & $0.096_{2.4}$ & $0.035_{1.0}$ & $0.138_{2.7}$ & $0.070_{1.7}$ & $0.305_{3.6}$ & $0.593_{7.7}$ \\
Combined: PC + SE & $0.073_{2.0}$ & $\mathbf{0.029_{0.2}}$ & $\mathbf{0.114_{2.3}}$ & $\mathbf{0.061_{1.6}}$ & $0.271_{3.6}$ & $0.532_{6.4}$ \\
Combined: PC + SE Probe & $0.069_{1.6}$ & $0.032_{0.5}$ & $0.129_{2.5}$ & $0.069_{1.6}$ & $0.302_{3.6}$ & $0.445_{4.3}$ \\
\addlinespace
\bottomrule
\end{tabular}
\end{table*}

\clearpage
\section{Combiner Ablation}
\label{sec:app-combiner-ablation}

Table~\ref{tab:app_combiner_ablation} compares Logistic 
Regression (LR) and a shallow MLP as combiners for fusing correctness and entropy signals. Across models and datasets, neither combiner consistently outperforms the other: LR wins or ties in the majority of cases on MedicalQA, while the MLP shows marginal gains on some TriviaQA and BioASQ configurations. Differences are small throughout and often fall within bootstrap confidence intervals. Given the absence of a consistent advantage for the MLP and the additional hyperparameter sensitivity it introduces, we adopt LR as our combiner throughout the main experiments.

\begin{table*}[h]

    \centering
    \caption{Ablation study comparing the classification performance (AUROC) of Logistic Regression (LR) and Multi-Layer Perceptron (MLP) combiners for fusing correctness and entropy signals.}
    \label{tab:app_combiner_ablation}
\begin{tabular}{l|cc|cc|cc}
\toprule
\textbf{Model} & \multicolumn{2}{c|}{\textbf{NLL}} & \multicolumn{2}{c|}{\textbf{SE}} & \multicolumn{2}{c}{\textbf{SE Probe}} \\
& \scriptsize{LR} & \scriptsize{MLP} & \scriptsize{LR} & \scriptsize{MLP} & \scriptsize{LR} & \scriptsize{MLP} \\
\midrule
Llama 3.2 3B (\textit{TriviaQA}) & $0.826_{3.4}$ & $\mathbf{0.831_{3.3}}$ & $0.863_{2.9}$ & $\mathbf{0.864_{2.8}}$ & $\mathbf{0.836_{3.2}}$ & $\mathbf{0.836_{3.2}}$ \\
Llama 3.2 3B (\textit{BioASQ}) & $\mathbf{0.674_{3.6}}$ & $0.674_{3.6}$ & $0.702_{3.5}$ & $\mathbf{0.707_{3.5}}$ & $\mathbf{0.693_{3.6}}$ & $0.693_{3.6}$ \\
Llama 3.2 3B (\textit{MedicalQA}) & $\mathbf{0.649_{5.2}}$ & $0.560_{6.2}$ & $\mathbf{0.666_{5.2}}$ & $0.639_{5.3}$ & $\mathbf{0.654_{5.4}}$ & $0.650_{5.5}$ \\
\midrule
Qwen3 4B (\textit{TriviaQA}) & $0.888_{2.6}$ & $\mathbf{0.891_{2.5}}$ & $0.910_{2.1}$ & $\mathbf{0.911_{2.1}}$ & $\mathbf{0.894_{2.4}}$ & $0.889_{2.5}$ \\
Qwen3 4B (\textit{BioASQ}) & $0.711_{4.3}$ & $\mathbf{0.721_{4.3}}$ & $\mathbf{0.722_{4.1}}$ & $0.721_{4.2}$ & $\mathbf{0.707_{4.3}}$ & $0.694_{4.4}$ \\
Qwen3 4B (\textit{MedicalQA}) & $\mathbf{0.667_{5.1}}$ & $0.608_{5.4}$ & $\mathbf{0.681_{4.8}}$ & $0.671_{4.7}$ & $\mathbf{0.680_{4.8}}$ & $0.669_{4.8}$ \\
\midrule
Gemma 3 4B (\textit{TriviaQA}) & $\mathbf{0.756_{4.5}}$ & $0.749_{4.5}$ & $0.785_{4.2}$ & $\mathbf{0.787_{4.1}}$ & $\mathbf{0.776_{4.2}}$ & $0.771_{4.3}$ \\
Gemma 3 4B (\textit{BioASQ}) & $\mathbf{0.751_{4.4}}$ & $0.741_{4.7}$ & $0.761_{4.3}$ & $\mathbf{0.761_{4.3}}$ & $0.753_{4.4}$ & $\mathbf{0.761_{4.2}}$ \\
Gemma 3 4B (\textit{MedicalQA}) & $\mathbf{0.770_{5.4}}$ & $0.736_{6.1}$ & $\mathbf{0.789_{5.0}}$ & $0.741_{5.5}$ & $\mathbf{0.749_{6.1}}$ & $0.745_{5.8}$ \\
\midrule
Ministral 8B (\textit{TriviaQA}) & $0.753_{4.3}$ & $\mathbf{0.760_{4.2}}$ & $0.796_{3.8}$ & $\mathbf{0.800_{3.7}}$ & $\mathbf{0.788_{3.7}}$ & $0.784_{3.7}$ \\
Ministral 8B (\textit{BioASQ}) & $0.767_{3.8}$ & $\mathbf{0.772_{3.8}}$ & $0.791_{3.6}$ & $\mathbf{0.795_{3.6}}$ & $\mathbf{0.774_{3.7}}$ & $0.771_{3.7}$ \\
Ministral 8B (\textit{MedicalQA}) & $\mathbf{0.560_{5.3}}$ & $0.548_{5.6}$ & $\mathbf{0.628_{4.9}}$ & $0.581_{5.0}$ & $\mathbf{0.565_{5.2}}$ & $0.546_{5.5}$ \\
\midrule
\bottomrule
\end{tabular}
\end{table*}

\clearpage
\section{Risk--Coverage Curves by Model: TriviaQA}
\label{sec:app-rc-curves}

\begin{figure*}[h]
    \centering
    \begin{subfigure}[t]{0.48\textwidth}
        \centering
        \includegraphics[width=\textwidth]{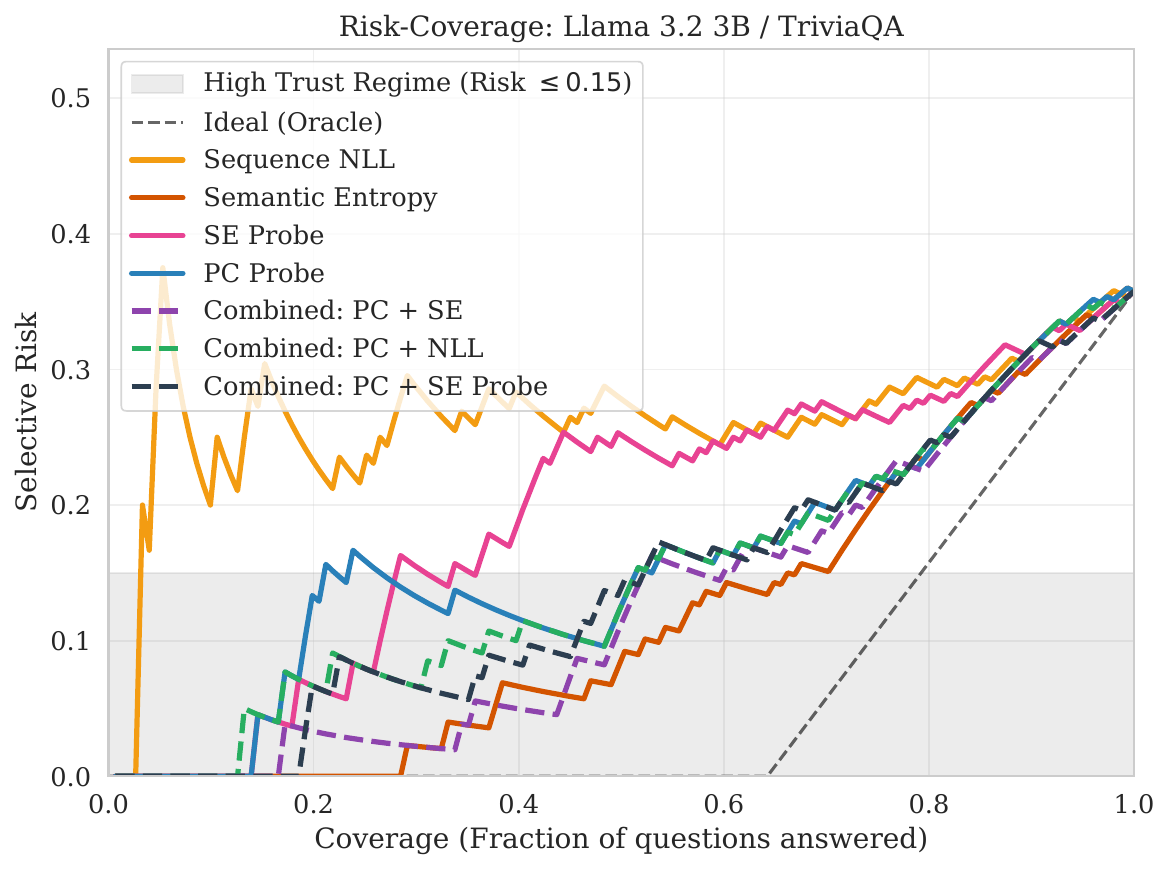}
        \caption{\textbf{Llama 3.2 3B.} Semantic Entropy is the strongest individual method, entering the high-trust regime at non-trivial coverage without requiring combination with the PC probe.}
        \label{fig:rc-llama}
    \end{subfigure}
    \hfill
    \begin{subfigure}[t]{0.48\textwidth}
        \centering
        \includegraphics[width=\textwidth]{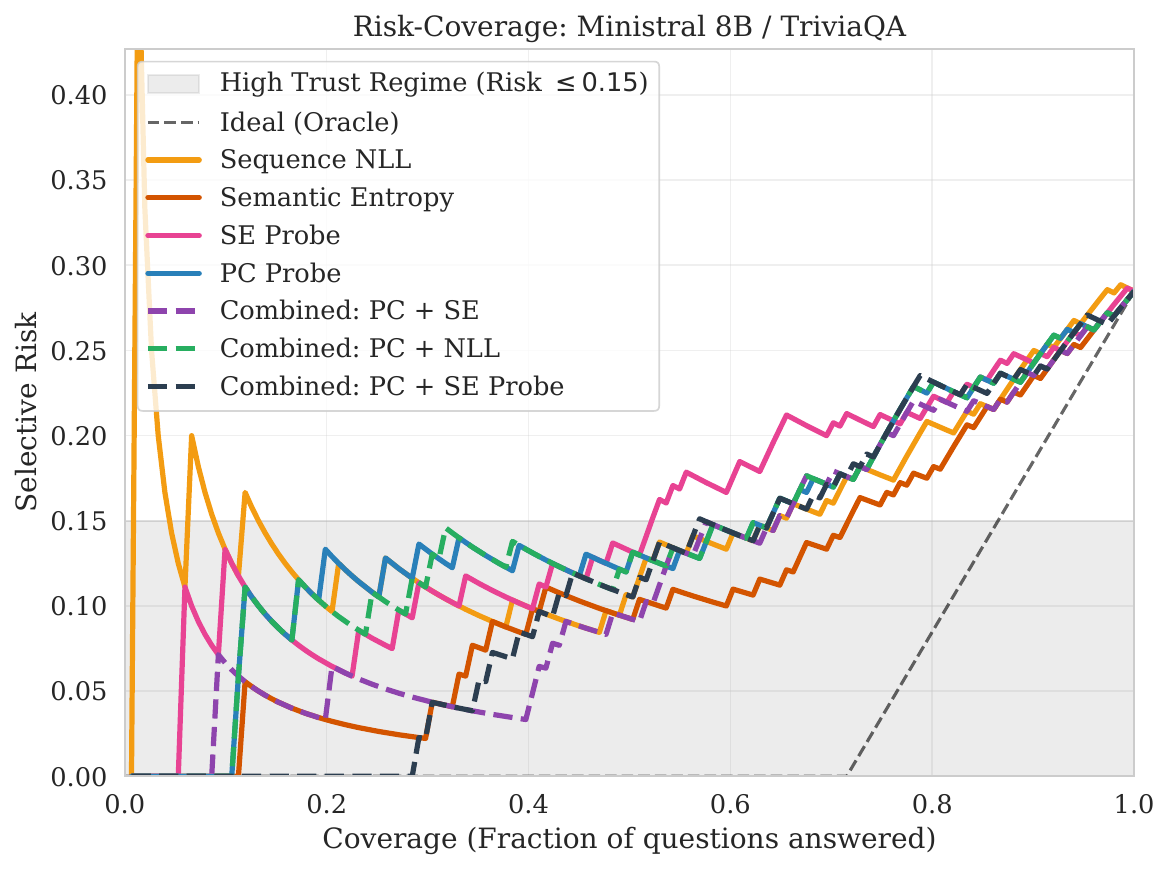}
        \caption{\textbf{Ministral 8B.} The combined PC + SE probe method achieves the best risk--coverage trade-off; entropy-only methods exhibit a higher risk floor in the high-trust regime.}
        \label{fig:rc-ministral}
    \end{subfigure}

    \vspace{1em}

    \begin{subfigure}[t]{0.48\textwidth}
        \centering
        \includegraphics[width=\textwidth]{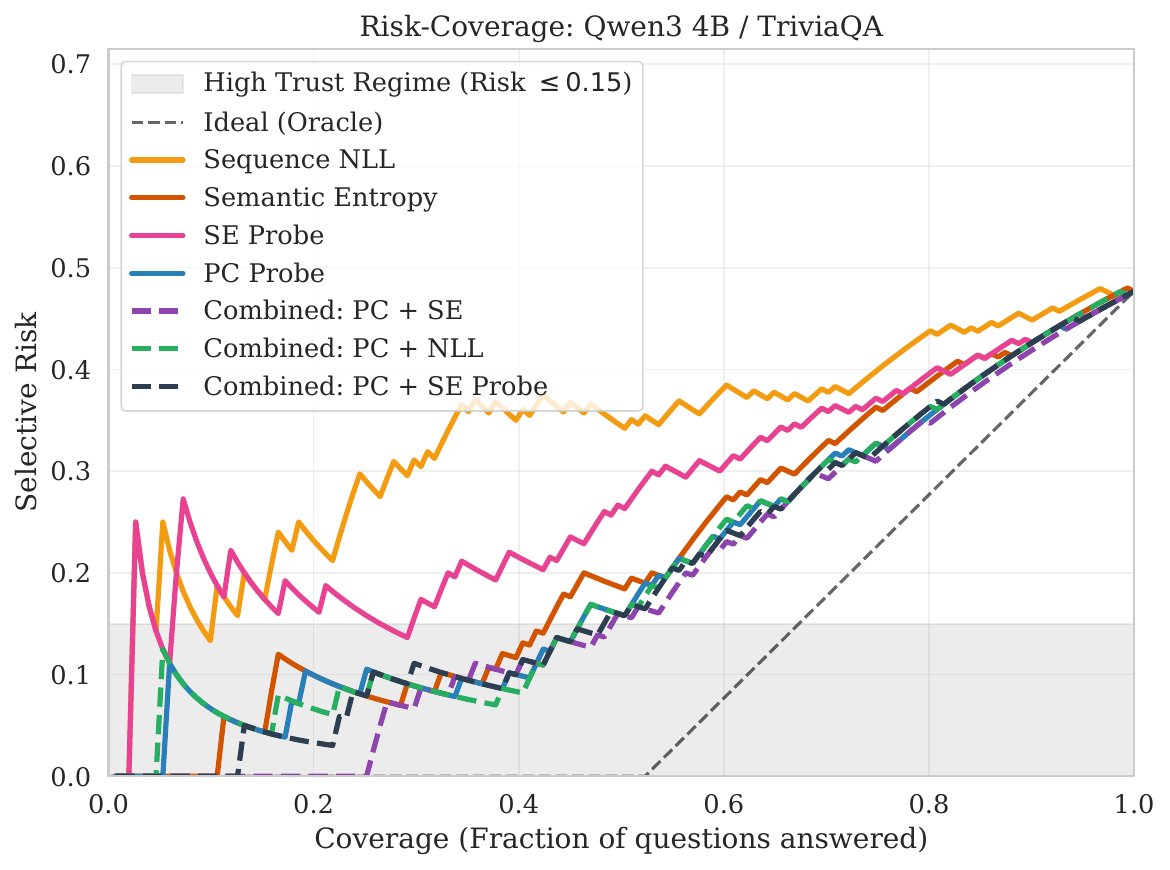}
        \caption{\textbf{Qwen3 4B.} The combined PC + SE method is able to lower selective risk at higher coverage compared to SE alone.}
        \label{fig:rc-qwen}
    \end{subfigure}
    \hfill
    \begin{subfigure}[t]{0.48\textwidth}
        \centering
        \includegraphics[width=\textwidth]{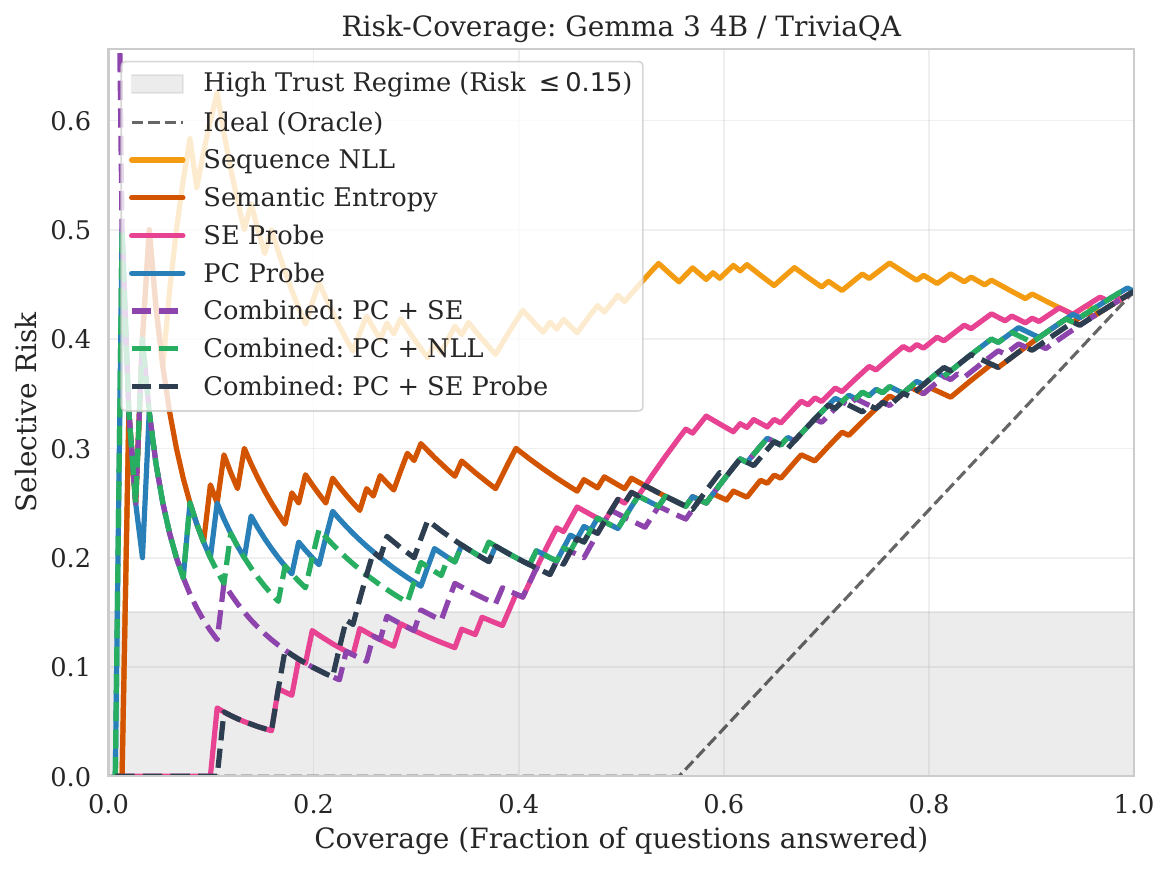}
        \caption{\textbf{Gemma 3 4B.} The SE Probe and combined methods perform similarly at low selective risk, but the combined method is optimal by overall E-AURC (Table \ref{tab:sp_results}).}
        \label{fig:rc-gemma}
    \end{subfigure}

    \caption{Risk--coverage curves for all four model families on TriviaQA. The shaded region marks the high-trust regime ($\alpha \leq 0.15$). The relative ranking of methods varies substantially across models, motivating model-aware evaluation of uncertainty quantification approaches.}
    \label{fig:app-rc-trivia}
\end{figure*}

Figure~\ref{fig:app-rc-trivia} shows risk--coverage curves for all four models on TriviaQA, illustrating how the relative effectiveness of entropy-based and combined methods varies substantially across model families.

\end{document}